\documentclass[11pt,a4paper]{article}
\usepackage[hyperref]{AACL-IJCNLP2020}
\usepackage{times}

\usepackage{latexsym}

\usepackage[whole]{bxcjkjatype} 

\usepackage{microtype}

\aclfinalcopy 
\def\aclpaperid{12} 

\title{Inference-only sub-character decomposition improves translation of unseen logographic characters}

\author{Danielle Saunders$^{\dag\ddagger}$ \and Weston Feely\thanks{ \hspace*{0.5em}Now at Amazon, work performed while at SDL} \and Bill Byrne$^{\dag}$ \\
    $^\dag$Department of Engineering, University of Cambridge, UK  \\
    $^\ddagger$SDL Research\\
      {\tt ds636@cam.ac.uk\hspace*{0.5em} wesfeely@gmail.com
     \hspace*{0.5em} wjb31@cam.ac.uk}}

\begin{document}

\maketitle
\begin{abstract}

Neural Machine Translation (NMT) on logographic source languages struggles when translating `unseen' characters, which never appear in the training data.  One possible approach to this problem uses sub-character decomposition for training and test sentences. However, this approach involves complete retraining, and its effectiveness for unseen character translation to non-logographic languages has not been fully explored.

We investigate existing ideograph-based sub-character decomposition approaches for Chinese-to-English and Japanese-to-English NMT, for both high-resource and low-resource domains. For each  language pair and domain we construct a test set where all source sentences contain at least one unseen logographic character. We find that complete sub-character decomposition often harms unseen character translation, and gives inconsistent results generally. 

We offer a simple alternative based on decomposition before inference for unseen characters only. Our approach allows flexible application, achieving translation adequacy improvements and requiring no additional models or training.
\end{abstract}

\section{Introduction}
While Neural Machine Translation (NMT) has evolved rapidly in recent years, not all of its successful techniques are equally applicable to all language pairs. A particular example is the representation and translation of unseen tokens, which do not appear in the training data. With techniques like  subword decomposition \citep{sennrich-etal-2016-neural}, an unseen word in an alphabetic language can in the worst case be represented as a sequence of characters. Since alphabetic languages usually have few unique characters, it is reasonable to assume that all of these `backoff' characters will be present in the limited model vocabulary. 

We focus instead on the translation of unseen Chinese and Japanese logographic characters into alphabetic languages, a task that remains a  challenge for NMT. Logographic writing systems may have many thousands of logograms, each representing at least one word, morpheme or concept as well as conveying phonetic and prosodic information. Inevitably some characters will either not be present in the training data, or will be present but too rare to be included in the vocabulary.

If the model is required to translate a previously unseen character, it will usually be replaced with an \texttt{UNK} (unknown word) token. The most likely outcome is that it will be ignored by the translation model, which will instead rely on the context of the unseen character to produce the translation. In the worst case, the presence of a previously-unseen character at inference time may  harm the translation quality. This is a particular concern for NMT in low-resource domains, when a model is less able to rely on lexical context. 

Many logographic characters share sub-character components\footnote{214 Kangxi Radicals are defined as a block in Unicode as of version 3.0 \citep{unicode2000unicode}. In this paper we follow prior work in using shallower decompositions which can include non-radical sub-character units.}, which can carry semantic or phonetic meaning (Table \ref{tab:examples}). An intuitive approach to the logogram sparsity problem in NLP uses sub-character decompositions in place of characters. Sub-character work in NMT has focused on the use of shared sub-characters to improve Chinese-Japanese translation \cite{zhang-komachi-2018-neural, zhang2019chinese}. In this approach all logograms are decomposed, and subword vocabularies are learned over sub-character sequences. 

\begin{table}[ht]
    \centering
    \small
    \begin{tabular}{|c|c|c|c|}
    \hline
    Char & Meaning& Sub-chars  & Semantic sub-char \\
    \hline
     森 & Forest &木 木 木  & 木 (Tree)\\
      鰯 & Sardine & 魚弱   & 魚 (Fish) \\
       校  & School &   木  交  & 木 (Tree) \\
\hline
    \end{tabular}
    \caption{Some characters with  sub-character decompositions given by CHISE. Not all decompositions or sub-characters convey the character's semantic meaning.}
    \label{tab:examples}
\end{table}
We identify two motivations for using sub-characters in logographic NMT:

\begin{enumerate}
    \item Sharing vocabularies between languages with similar sub-character decompositions, as in Chinese-Japanese translation.
    \item Representing unseen characters -- those not appearing in the training data -- in semantically meaningful ways.
\end{enumerate}

Our hypothesis is that, while complete sub-character decomposition for \emph{all} characters might be useful in case 1, only \emph{some} characters benefit from only semantic elements of the decomposition in case 2. The focus of this work is case 2. Our contributions are as follows: 
\begin{itemize}
    \item We compare ideograph-based sub-character schemes for Chinese-to-English and Japanese-to-English NMT with a strong BPE subword baseline, for both high- and low-resource domain translation. 
    \item We evaluate both on general test sets, and on challenge sets which we construct such that all sentences have at least one character that was not seen in the training data. To the best of our knowledge, this is the first attempt to analyze the impact of sub-character decomposition on unseen character translation.
    \item We demonstrate that, counter-intuitively, training models with indiscriminate sub-character decomposition can harm unseen character translation, and also gives inconsistent performance on sentences with no unseen characters. 
    \item We instead propose a set of extremely straightforward inference-time sub-character decomposition schemes requiring no additional models or training.
\end{itemize}

\subsection{Related work}

In NMT, applying radical decomposition before learning a Byte Pair Encoding (BPE, 
\citet{sennrich-etal-2016-neural}) vocabulary has been shown to improve Chinese-Japanese supervised and unsupervised translation over a standard character BPE representation \cite{zhang-komachi-2018-neural, zhang2019chinese}. However, Chinese-Japanese translation benefits from a high proportion of shared sub-characters.  The impact of sub-character decomposition for translating unseen logographic characters to an alphabetic language that cannot share  sub-characters has not been fully explored.   

\citet{zhang-komachi-2018-neural} also explore translation from Chinese and Japanese with sub-character decomposition into English. However, they translate to word-based English sentences instead of a stronger BPE representation, and they do not assess the effect of sub-character decomposition on unseen characters. \citet{kuang2018apply} likewise train NMT models with factored  sub-character information for Chinese-to-English translation but use words instead of BPE units as their baseline decomposition for both languages. 

\citet{zhang2019subcharacter} do explore sub-character decomposition for BPE-based Chinese-to-English NMT. They find that training with sub-character decomposition  alone does not give quality improvements in this case, although it has the practical advantage of a smaller vocabulary size. Our findings echo these, but we confirm them for Japanese-to-English translation and translation of unseen characters specifically. 

Outside of translation, use of sub-character decomposition has been shown to improve learning of character embeddings for Chinese \cite{sun2014radical} and language modelling for Japanese \cite{nguyen-etal-2017-sub}. Sub-character decomposition has also been applied to sentiment classification \cite{ke2017radical}, text classification \citep{toyama-etal-2017-utilizing} and word similarity tasks, the last with mixed results \cite{karpinska-etal-2018-subcharacter}. 

We consider work on NMT with byte-level subwords  \citep{costa-jussa-etal-2017-byte,  wang2019neural} as complementary to this work. Representing text at the level of bytes  allows any logographic character with a Unicode representation to be included in the model vocabulary. However, inclusion in the vocabulary does not guarantee that the model learns a good character representation, and such schemes do not leverage the semantic or phonetic information available in sub-character decompositions of unseen characters.

\section{Sub-character decomposition}

Over 80\% of Chinese characters can be broken down into both a semantic and a phonetic component \citep{liu2010holistic}. The semantic meaning of a Chinese character often corresponds to the sub-character occupying its top or left position \citep{hoosain1991psycholinguistic}. These may be -- but are not always -- radicals: sub-characters that cannot be broken down any further. However, radicals in these positions are not necessarily directly meaningful. For example, radical 魚 (`fish') has a clear semantic relationship with the character 鰯 (`sardine'), but the semantic connection of radical  木 (`tree') to  character 校 (`school') is more abstract. Example decompositions are given in Table \ref{tab:examples}. The phonetic component is less likely to be helpful for translation to a non-logographic language, except in the case of transliterations.  

\subsection{Training with sub-character decomposition}

We first explore the impact of two variations on ideograph-based sub-character decomposition applied to all characters in the source language. Following  \citet{zhang2019chinese} we use decomposition information from the CHISE project\footnote{Accessed via https://github.com/cjkvi/cjkvi-ids}, which provides ideograph sequences for CJK (Chinese-Japanese-Korean) characters. As well as ideographs, the sequences include ideographic description characters (IDCs), which convey the structure of an ideograph. While  \citet{zhang2019chinese} use IDC information for Chinese-Japanese translation, use of structural sub-character information has not yet been explored for NMT to an alphabetic language.

IDCs may convey useful information about which sub-character component is likely to be the semantic or phonetic component, but they also make character representations significantly longer. We therefore compare training with sub-character decompositions with and without the IDCs. 

\subsection{Inference-only sub-character decomposition}
\label{decomposition}
Applying sub-character decomposition to all characters for training decreases the vocabulary size, but significantly lengthens source sequences. Additionally, these schemes apply decomposition to all source characters, regardless of whether they benefit from decomposition. We propose an alternative approach which applies sub-character decomposition only to unseen characters at inference time. 

We apply decomposition if a test source sentence 1) contains an unseen character which 2) can be decomposed into at least one sub-character that is already present in the vocabulary. We do not include the entire decomposition, but keep only the sub-characters already in the model vocabulary.  We experiment with both keeping all in-vocabulary sub-characters, and keeping only the leftmost in-vocabulary sub-character, which is frequently the semantic component. We consider the left-only approach to be a reasonable heuristic since in many cases other components do not contribute semantic meaning.

The inference-only decomposition  approach has several advantages over training with sub-character decomposition. It is extremely fast, since decomposition is a pre-processing step before inference. It does not require training from scratch with very long sequences, which can harm overall performance. Sentences without unseen characters, which are unlikely to benefit from decomposition, are left completely unchanged by the scheme.

Finally, the scheme is very flexible: decomposition can be applied to individual unseen characters on a case-by-case basis if necessary. For example, the presence of the 魚 (`fish') radical on the left of a character very often indicates that the character is for a type of fish, so applying inference-only decomposition to such characters will improve adequacy. Characters can in principle be excluded from decomposition if they do not benefit from it.

We convert some sub-character components to their base forms to improve character coverage. A small number of components change form in some cases. For example, 水 (`water'), can exist as its own character or as a radical, but often becomes 氵when used on the left hand side of a character (e.g. 池, `pond'). We manually define 30 such cases for inference-only decomposition, swapping the changed radical (unlikely to be in the vocabulary) for its base form (often in the vocabulary). This is unneeded when training with sub-character decomposition as all forms can be included in the vocabulary.

Even when radicals are replaced with their base form, not all radicals will be present in the vocabulary of a non-sub-character model. To address this problem we propose replacing the out-of-vocabulary radical with an in-vocabulary, non-radical character that conveys a related semantic meaning. Experimentally, we attempt this with a single character for both Chinese and Japanese,  replacing radical 疒 (`illness'), which is not in the vocabulary, with character `病' (`illness').

Finally, a very simple approach to unseen sub-characters is to  remove them from source sentences. This makes it  unlikely that the character will be correctly translated, but saves the model from translating an \texttt{UNK}. We only apply this to characters which could be decomposed, so  \texttt{UNK} may still occur.

Examples of real sub-character decompositions for all schemes in this work are shown in  Table \ref{tab:example-decomp}.

\begin{table}[ht]
    \centering
    \small
    \begin{tabular}{|p{4.2cm}|c|c|}
    \hline
    Decomposition & 鰯  & 瘡  \\
    \hline
    Baseline & \texttt{UNK} & \texttt{UNK}\\
    Training decompose  & 魚弱   & 疒倉  \\
    Training decompose (IDC)    & ⿰ 魚弱  & ⿸疒倉 \\
    \hline
    Inference-only remove  & & \\
    Inference-only decompose & 魚弱 & 倉\\
    Inference-only decompose (left) & 魚 & 倉 \\
    Inference-only decompose (replace unseen radical) &魚弱 & 病倉 \\
    Inference-only decompose (left, replace unseen radical) &魚 & 病 \\

\hline
    \end{tabular}
    \caption{Training and inference-only decompositions used in this work. Example representations for two characters: 鰯 (`sardine', semantic component 魚 `fish') and  瘡 (`sores', semantic component 疒 `illness'). Sub-character 疒 is not in the vocabulary, so does not appear in inference-only decomposition unless swapped with an in-vocabulary character e.g. 病 (`illness').}
    \label{tab:example-decomp}
\end{table}
\begin{table*}[ht]
    \centering
    \small
    \begin{tabular}{|l|cc|cc|}
    \hline
       \textbf{Set}  & \multicolumn{2}{|c|}{\textbf{Chinese-English}} & \multicolumn{2}{|c|}{\textbf{Japanese-English}}  \\
        \hline
    Training data source    & Proprietary & CAS & ASPEC & KFTT \\
    \hline
    Train & 50M & 3M &  2M & 330K\\
    General test set & 2000  & 3981 & 1812 & 1160  \\
    Unseen chars test set & 2140 &  1360 & 336 & 2243\\
    \hline
    \end{tabular}
    \caption{Sentence counts for Chinese-English and Japanese-English training and test sets.  Chinese-English proprietary and CAS training corpora have no standard test sets, so we use the WMT news task WMT19 and WMT18 test sets respectively. The `unseen chars' test sets are held out from the corresponding training sets such that every sentence has at least one unseen decomposable logographic character.}
    \label{tab:my_label}
\end{table*}

\section{Experiments}

\begin{table*}[ht]
    \centering
    \small
    \begin{tabular}{|p{2.3cm}|cc|cc|cc|cc|}
    \hline
       \textbf{Decomposition}  & \multicolumn{4}{|c|}{\textbf{Chinese-English}} & \multicolumn{4}{|c|}{\textbf{Japanese-English}}  \\
    \textbf{(training)} & \multicolumn{2}{|c|}{Higher-resource} & \multicolumn{2}{|c|}{Lower-resource}  & \multicolumn{2}{|c|}{Higher-resource} & \multicolumn{2}{|c|}{Lower-resource}   \\

        & WMT19  & Unseen & WMT18 & Unseen &  ASPEC & Unseen & KFTT & Unseen \\
        \hline
    None (Baseline)  & \textbf{25.2} & \textbf{22.6}&  \textbf{18.3}& \textbf{12.4}& \textbf{28.3} & 13.5& \textbf{16.9}&\textbf{13.3} \\ 
    Decompose  & 24.9 & \textbf{22.6}& 17.5& 11.4& 26.9& \textbf{14.8} & 16.2& 12.5\\ 
    Decompose IDC & 24.8 & 22.5 & 18.2 &\textbf{12.4} & 26.4 & 14.7 & 16.2& 12.4\\
    \hline
    \end{tabular}
    \caption{BLEU scores for training with different decomposition schemes for higher- and lower-resource test sets. Baseline has no sub-character decomposition. Sub-character decomposition during training fails to  improve  general translation, and only improves unseen set translation for ASPEC.}
    \label{results-train-decomp}
\end{table*}

\begin{table*}[ht]
    \centering
    \small
    \begin{tabular}{|p{3.5cm}|cc|cc|cc|cc|}
    \hline
       \textbf{Decomposition  }  & \multicolumn{4}{|c|}{\textbf{Chinese-English}} & \multicolumn{4}{|c|}{\textbf{Japanese-English}}  \\
    \textbf{(inference)} & \multicolumn{2}{|c|}{\textbf{Higher-resource}} & \multicolumn{2}{|c|}{\textbf{Lower-resource}} & \multicolumn{2}{|c|}{\textbf{Higher-resource}} & \multicolumn{2}{|c|}{\textbf{Lower-resource}}  \\

        & WMT19  & Unseen & WMT18 & Unseen &  ASPEC & Unseen & KFTT & Unseen \\
                \hline
    None (Baseline) & 25.2 & \textbf{22.6}&  18.3& \textbf{12.4}& 28.3 & 13.5& \textbf{16.9}&13.3 \\ 
    Remove unseen  & 25.2 & \textbf{23.0} & 18.3 &   \textbf{12.4}&\textbf{28.4} &\textbf{15.0}  &  16.8 & \textbf{13.5} \\ 
    Decompose unseen &25.2 & 22.7 & 18.3& 11.8 & 28.3& 14.2& 16.7 & 12.9\\
    Decompose unseen (left) & 25.2& \textbf{23.0} &18.3 &12.0 &\textbf{28.4} & 14.6 & 16.7 & 13.3\\
    \hline
    \end{tabular}
    \caption{Higher- and lower-resource test set BLEU scores for the baseline models of Table  \ref{results-train-decomp} with different inference-time decomposition methods. Line 1 is duplicated from Table \ref{results-train-decomp}. Inference-time decomposition matches the baseline on general test sets, and some unseen sets see  BLEU improvement.}
    \label{results-infer-decomp}
\end{table*}

\begin{table*}[ht]
\centering
\small
\begin{tabular}{|p{3.7cm}|p{11.8cm}|}
\hline
Japanese source (ASPEC) & 空気 中 で は [鰯]油 が 最も 酸化 さ れ やす く ， ついで 亜麻 仁 油 ， 大豆 油 の 順 で あ っ た。\\
English reference & Due to its high contents of DHA and EPA, [sardine] oil FFA was most rapidly oxidized in air, followed by linseed and soybean oil FFAs. \\
\hline
Baseline &  In the air, the sate oil was most easily oxidized, followed by linseed oil and soybean oil.\\
Training decompose &  In air, salmon oil was most susceptible to oxidation, followed by linseed oil and soybean oil. \\
Inference decompose (left)& Fish oil was most oxidized in air, followed by linseed oil and soybean oil.\\
\hline
Japanese source (KFTT) & 康元 元年 ( 1256 年 ) 赤 斑 [瘡]に よ り 死去 。 \\
English reference & In 1256, he died from [measles].\\
\hline
Baseline & In 1256, he died of a red spot.\\
Training decompose &  In 1256, he died of a spear. \\
Inference decompose (left)& In 1256, he died from a red spot storehouse.\\
Inference decompose (left, replace radical)& In 1256, he died from a red spot disease.\\

\hline
\end{tabular}
\caption{Examples of translation with different decomposition schemes from the Japanese-English `unseen' sets. We compare the most consistent training decomposition (no IDCs) and inference-only decomposition (left-only) to the baseline. In the second  example, we additionally compare swapping the unseen radical with an in-vocabulary character. Unseen characters and (approximate) reference translations are marked in square brackets.}\label{tab:textexamples}
\end{table*}

\subsection{Datasets}
For both Chinese-English and Japanese-English, we first train a baseline model on a larger corpus and then adapt the same model to a smaller corpus. This lets us evaluate unseen character translation in both higher- and lower-resource settings. In both cases we evaluate on a corresponding standard test set where available, as well as an unseen characters test set. The latter is constructed from training sentences containing at least one decomposable logographic character otherwise not appearing in the training set.  These sentences are held out from the the training data, so any logographic characters appearing only in an  `unseen chars' set are not seen at all during training. 

To construct the unseen character set for the higher-resource domain we hold out training sentences with logographic characters appearing infrequently\footnote{No more than two occurrences for Chinese or three for Japanese} in the whole corpus, then filter for source/target sentence length ratio less than 3.5. We build the BPE vocabularies \citep{sennrich-etal-2016-neural}  on the high-resource domain training set. The baseline source and all target BPE vocabularies consist of character sequences, while the sub-character BPE vocabularies consist of sub-character sequences, following \citet{zhang-komachi-2018-neural}. For the lower-resource domains the unseen sets are held-out sentences containing logographic characters not in the baseline source vocabulary, filtered as before.

For Chinese-English our baseline model is trained on a proprietary parallel training data set containing web-crawled data from a mix of domains. We learn separate Chinese and English BPE vocabularies on this corpus with 50K merges. For the lower-resource-domain model we adapt to 3M sentence pairs from  publicly available corpora made available by the Chinese Academy of Sciences (CAS)\footnote{Casia2015 and Casict2015 corpora from  http://nlp.nju.edu.cn/cwmt-wmt/}. Since neither of these training sets have standard test set splits, we use the WMT news task test sets WMT19 and WMT18 zh-en for general evaluation of the higher- and lower-resource cases respectively \citep{barrault-etal-2019-findings}. WMT19 contains only seen characters, as do all but 2 lines of WMT18.

For Japanese-English, we train the higher-resource model on 2M scientific domain sentence pairs from the ASPEC corpus \citep{aspec}. We learn separate Japanese and English BPE vocabularies on this corpus with 30K merges. Our smaller domain is the Kyoto Free Translation Task (KFTT) corpus \citep{neubig11kftt}. We use the standard test sets for general evaluation. In the ASPEC test set 36 (2\%) sentences contain unseen decomposable characters, as well as 180 (15.5\%) sentences in the KFTT test set.  

\subsection{Experimental setup and evaluation}
Our NMT models are all Transformer models \citep{vaswani2017attention}. We use 512
hidden units, 6 hidden layers, 8 heads, and a batch
size of 4096 tokens in all cases. We train for 300K steps for the Chinese-English and for 240K steps for the Japanese-English higher-resource domain models. For the lower resource domains we fine-tune the trained models for 30K and 10K steps respectively.

We conduct inference via beam search with beam size 4. For ASPEC evaluation we evaluate Moses tokenized English with the multi-bleu tool to correspond to the official WAT evaluation. For all other results we report detokenized English using the SacreBLEU tool\footnote{BLEU+case.mixed+numrefs.1+smooth.exp+\\tok.13a+version.1.4.8} \citep{post-2018-call}. All BLEU is for truecased English.

\subsection{Results}
We have two requirements when using sub-character decomposition for unseen character translation:
\begin{itemize}
    \item Sets with few unseen characters (all general test sets except KFTT) should not experience performance degradation in terms of BLEU. 
    \item Translation performance on unseen characters should improve. 
\end{itemize} 

Unseen character translation improvement may not be detectable by BLEU score, since the unseen character sets may only have one or two unseen characters per sentence. Moreover generating a hypernym, such as `fish' instead of `sardine' for 鰯, would not improve BLEU, despite being a more adequate translation than \texttt{UNK} and a more correct translation than e.g. `salmon'. Consequently we also give examples for the most promising schemes.
\subsubsection{Training with decomposition}

In Table \ref{results-train-decomp} we give results after training with sub-character decomposition schemes. We compare decomposition with and without structural information (IDCs) to a strong BPE baseline. On general test sets, we see BLEU degradation compared to the baseline, especially for Japanese-English. We note that our Japanese-English ASPEC decomposed-training score is similar to the result for the same set achieved by \citet{zhang-komachi-2018-neural} with ideograph decomposition. However, our non-decomposed baseline is much stronger, and so we are not able to replicate their finding that training with sub-character decomposition is beneficial to NMT from logographic languages to English. We suggest this degradation may be the result of training and inference with much longer sequences, which are well-established as challenging for NMT \citep{koehn-knowles-2017-six}.  

Interestingly we find that adding IDCs, which lengthen sequences, performs slightly better for the lower-resource than for higher-resource cases, especially for Chinese-English. A possible explanation is that the longer sequences regularize adaptation in these cases, avoiding overfitting to the highly specific  lower-resource domains. However, these cases still show degradation relative to the baseline.

On the unseen sets, training with sub-character decomposition outperforms the baseline in terms of BLEU for the ASPEC unseen set. However, this is not a consistent result, with the baseline performing best or joint best in all other cases.
\subsubsection{Inference-only decomposition}

Table \ref{results-infer-decomp} gives results for our inference-only unseen character decomposition schemes, compared to the baseline with no decomposition. Inference time decomposition has no effect on the Chinese-English test sets with no unseen characters. This is as we expect, since these test sets are unchanged. For Japanese-English a slight decrease on the KFTT general set (about 15\% sentences with unseen characters) is balanced by a small improvement on the ASPEC general set (2\% sentences with unseen characters). These results are a strong advantage compared to training decomposition, which must be applied to all sentences whether they benefit or not, often degrading performance. 

Test sets with many unseen characters have a range of BLEU performance under inference-time decomposition. One consistent result is that left-only decomposition gives better scores than using all sub-characters. This may be explained by the fact that representing a character as multiple sub-characters may lead the model to generate a separate translation for each sub-character, harming performance. By contrast the leftmost sub-character tends to be the semantic component so may give good translation performance alone.

As a precision-based metric, BLEU is not an ideal measure of improving unseen character translation. Any such improvements under decomposition are more likely to improve adequacy  than precision, since they often involve introducing synonyms or hypernyms. This difficulty is highlighted by the strong performance of the `remove unseen' scheme which simply deletes unseen decomposable characters from source sentences. Clearly, such a scheme cannot improve the translation of these characters, although it may reduce the number of hypothesis tokens, inadvertently improving precision and therefore BLEU.

The higher performance of the decompose (left) scheme is more promising, since this is likely to actually generate translations for unseen characters. On a similar note, replacing the unseen  `illness' radical with a character conveying the same semantic meaning as described at the end of Sec. \ref{decomposition} does not affect BLEU for any set, but we do see noticeable improvements in adequacy for the handful of affected sentences.

\subsubsection{Qualitative evaluation}
We provide example translations under different training and inference decomposition schemes in Table \ref{tab:textexamples}. We observe some interesting differences in adequacy between training decomposition and inference-only decomposition.  In particular, both   translations with training decomposition feature a plausible but incorrect translation. With inference-only decomposition the translation is less fluent, but more generic and consequently more correct.

We note that  training with sub-character decomposition has an unfortunate tendency to translate over-specific terms from spurious sub-character matches. For example, in the first (ASPEC) Japanese example, 魚 (`fish') is also the radical in 鮭 (`salmon'), and in the second (KFTT) Japanese example,  倉 (`storehouse') is also a major component in 槍 (`spear'). The model trained with sub-character decompositions therefore produces `salmon' and `spear' instead of `sardine' and `measles'. Meanwhile the inference-only left-radical heuristic produces `fish' and `disease', both of which are  correct translations, if not reference-matching. 

We identify this pattern throughout the unseen-character sets for certain characters in particular. Characters for concrete nouns, such as types of fish, illness, bird, tree, and so on tend to be well-handled by  inference-only decomposition with the left-sub-character heuristic and failed by the training decomposition scheme.

More abstract characters are more challenging for both schemes, such as those with radical 心 (`heart') which often refer to an emotion. However, a major benefit of our approach is its flexibility; such poorly-handled characters could simply be excluded from the decomposition scheme, or replaced with a more appropriate non-radical character as we do for the `illness' radical 疒. Future work on this problem could involve determining the most relevant sub-character component of an character, if any, rather than the simple left-only heuristic.

\section{Conclusions}
We explore the effect of sub-character decomposition on NMT from logographic languages into English. During training decomposition may hurt general translation performance without necessarily helping unseen character translation. We propose a flexible inference-time sub-character decomposition procedure which targets unseen characters, and show that it aids adequacy and reduces misleading overtranslation in unseen character translation. The scheme is straightforward, requires no additional models or training, and has no negative impact on sentences without unseen characters.

\section*{Acknowledgments}
This work was supported by EPSRC grants EP/M508007/1 and EP/N509620/1. Some resources provided by the Cambridge Tier-2 system operated by the University of Cambridge Research Computing Service\footnote{\url{http://www.hpc.cam.ac.uk}} funded by EPSRC Tier-2 capital grant EP/P020259/1. Work carried out by D Saunders during a research placement at SDL plc.

\bibliography{anthology,aacl-ijcnlp2020}
\bibliographystyle{acl_natbib}

\end{document}